\documentclass[letterpaper, 10 pt, conference]{ieeeconf}
\IEEEoverridecommandlockouts    
\usepackage{graphics}           
\usepackage{times}              
\usepackage{amsmath}            
\usepackage{amssymb}            
\usepackage{graphicx}
\usepackage{algorithm}
\usepackage[noend]{algpseudocode}
\usepackage{booktabs}
\usepackage{color}
\definecolor{instructioncolor}{rgb}{.5,.5,.5}

\usepackage[font=small]{caption}

\def\secref#1{Sec.~\ref{#1}}
\def\figref#1{Fig.~\ref{#1}}
\def\tabref#1{Tab.~\ref{#1}}
\def\eqref#1{Eq.~(\ref{#1})}

\makeatletter
\usepackage{xspace}
\DeclareRobustCommand\onedot{\futurelet\@let@token\@onedot}
\def\@onedot{\ifx\@let@token.\else.\null\fi\xspace}

\def\etal{{et al}\onedot}
\makeatother

\usepackage{array}
\newcolumntype{L}[1]{>{\raggedright\let\newline\\\arraybackslash\hspace{0pt}}m{#1}}
\newcolumntype{C}[1]{>{\centering\let\newline\\\arraybackslash\hspace{0pt}}m{#1}}
\newcolumntype{R}[1]{>{\raggedleft\let\newline\\\arraybackslash\hspace{0pt}}m{#1}}

\renewcommand{\d}[1]{\b {#1}}

\newcommand{\m}[1]{{\mbox{{\sffamily\slshape{#1\/}}}}}

\newcommand{\mF}{\m F}

\newcommand{\mI}{\m I}

\usepackage{multirow}
\usepackage{hyperref}

\renewcommand{\d}[1]{\mbox{\boldmath$#1$}}	%
\title{\LARGE \bf BEVDiffLoc: End-to-End LiDAR Global Localization in BEV View based on Diffusion Model}

\author{Ziyue Wang \and Chenghao Shi* \and Neng Wang \and Qinghua Yu \and Xieyuanli Chen \and Huimin Lu
  \thanks{All the authors are with the College of Intelligence Science and Technology, and the National Key Laboratory of Equipment State Sensing and Smart Support, National University of Defense Technology, China.}%
  \thanks{* indicates the corresponding authors: C. Shi (shichenghao17@nudt.edu.cn)}
  \thanks{This work was supported in part by the National Science Foundation of China (Grant No. 62403478, U22A2059 and 62203460), Young Elite Scientists Sponsorship Program by CAST (No. 2023QNRC001), and  Major Project of Natural Science Foundation of Hunan Province (Grant No. 2021JC0004).
  }%
}

\begin{document}
\maketitle
\thispagestyle{empty}
\pagestyle{empty}

\begin{abstract}
Localization is one of the core parts of modern robotics. Classic localization methods typically follow the retrieve-then-register paradigm, achieving remarkable success. Recently, the emergence of end-to-end localization approaches has offered distinct advantages, including a streamlined system architecture and the elimination of the need to store extensive map data. Although these methods have demonstrated promising results, current end-to-end localization approaches still face limitations in robustness and accuracy.
Bird’s-Eye-View (BEV) image is one of the most widely adopted data representations in autonomous driving. It significantly reduces data complexity while preserving spatial structure and scale consistency, making it an ideal representation for localization tasks. However, research on BEV-based end-to-end localization remains notably insufficient. To fill this gap, we propose BEVDiffLoc, a novel framework that formulates LiDAR localization as a conditional generation of poses.
Leveraging the properties of BEV, we first introduce a specific data augmentation method to significantly enhance the diversity of input data. Then, the Maximum Feature Aggregation Module and Vision Transformer are employed to learn robust features while maintaining robustness against significant rotational view variations. Finally, we incorporate a diffusion model that iteratively refines the learned features to recover the absolute pose. Extensive experiments on the Oxford Radar RobotCar and NCLT datasets demonstrate that BEVDiffLoc outperforms the baseline methods. Our code is available at \href{https://github.com/nubot-nudt/BEVDiffLoc}{https://github.com/nubot-nudt/BEVDiffLoc}.
\end{abstract}

\begin{figure}[t]
\centering
\includegraphics[width=\columnwidth]{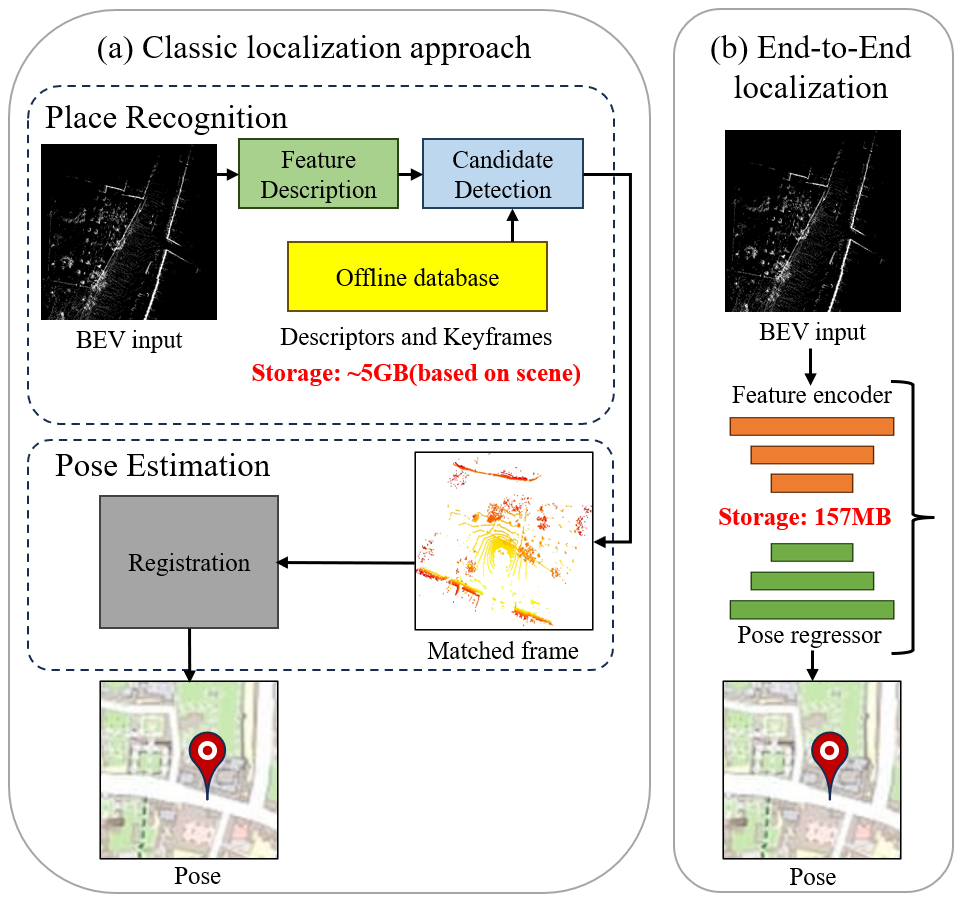}
\caption{Comparison between classic localization approaches and end-to-end localization methods. Classic localization approaches typically follow a two-step process: place recognition followed by pose estimation. This approach demands substantial memory resources to store map database. Furthermore, their reliance on multiple cascaded modules makes optimization challenging. In contrast, end-to-end methods implicitly encode environmental information into network parameters, resulting in a simpler structure and reduced storage requirements.}
\vspace{-0.5cm}
\label{fig:motivation}
\end{figure}

\section{Introduction}
\label{sec:intro}

LiDAR-based localization is one of the core technologies in modern robotics due to its unparalleled ability to provide accurate environmental geometric information. This paper focuses on the problem of LiDAR-based localization in the presence of prior environmental knowledge. Currently, mainstream approaches typically involve retrieving target point clouds from a prebuilt database and subsequently performing registration to determine the pose of the query~\cite{Goswami2024ral,kim2021tro,Uy2018cvpr,shi2024tro,luo2024bevplace++}. However, this strategy not only requires computationally intensive registration processes but also demands significant memory resources to store map points and descriptors. Recently, mapless localization systems have demonstrated significant potential in enabling end-to-end localization. These methods implicitly encode environmental information into network parameters through training, allowing for direct pose estimation based on current sensor observations. By storing only the network parameters, these approaches significantly reduce the need for memory-intensive maps and descriptors, as shown in~\figref{fig:motivation}.

The sparsity and non-uniformity of LiDAR point clouds are well-known challenges in the field. A widely adopted solution is to project point clouds into range images~\cite{chen2020rss,chen2021icra}. Such an approach yields promising results by reducing the complexity of point cloud data processing. However, due to the nature of spherical projection, range images are susceptible to scale distortion, which ultimately limits their localization accuracy.

Bird’s-Eye-View (BEV)-based point cloud representation has recently gained increasing attention in autonomous driving perception and localization. BEV significantly simplifies data complexity while clearly illustrating the position and size of objects. For localization tasks, compared to range images, BEV images provide more stable target scales and spatial relationships, as well as better generalization capabilities. However, during the process of generating BEV images, there may be point cloud stacking, which can result in information loss. Therefore, despite its advantages, according to recent study of Wang~\etal~\cite{wang2023cvpr}, research on BEV-based end-to-end LiDAR-based localization remains limited. This work aims to fill this gap.

In this paper, we propose BEVDiffLoc, a novel framework that achieves state-of-the-art end-to-end localization based on BEV representation. Our approach incorporates a Maximum Feature Aggregation (MFA) module to extract orientation-equivariant features, followed by a Vision Transformer (ViT)~\cite{Ando2023cvpr} to enhance the network’s understanding of spatial scene structures. Leveraging the orientation-equivariant property, the ViT can effectively handle significant rotational view changes, thereby better capturing positional information in the map. For pose regression, we adopt a diffusion model conditioned on the robust features, progressively refining the poses from noisy inputs.
To fully leverage the advantages of BEV representation, we introduce a specific data augmentation strategy for BEV images. This strategy involves first stitching BEV images into a map and then sampling new BEV images from random positions and random orientation, significantly enhancing the diversity of the input data.

Our contributions can be summarized as follows:
\begin{itemize}
\item We introduce BEVDiffLoc, a novel BEV-based end-to-end localization framework that achieves state-of-the-art performance among existing works.

\item We incorporate the Maximum Feature Aggregation (MFA) module and the Vision Transformer (ViT) module for feature extraction, enabling the network to robustly handle significant rotational view changes and demonstrating its effectiveness in end-to-end localization tasks.

\item To maximize the benefits of BEV representation, we concatenate BEV images and generate input data by randomly sampling positions from the map, enhancing training data diversity and thereby boosting localization performance.

\item We open-source our approach to benefit the research community.
\end{itemize}

We evaluated our approach on multiple datasets, demonstrating the leading performance of our method and validating the contributions of each proposed component to the overall performance improvement.

\begin{figure*}[htbp]
\centering
\includegraphics[width=\textwidth]{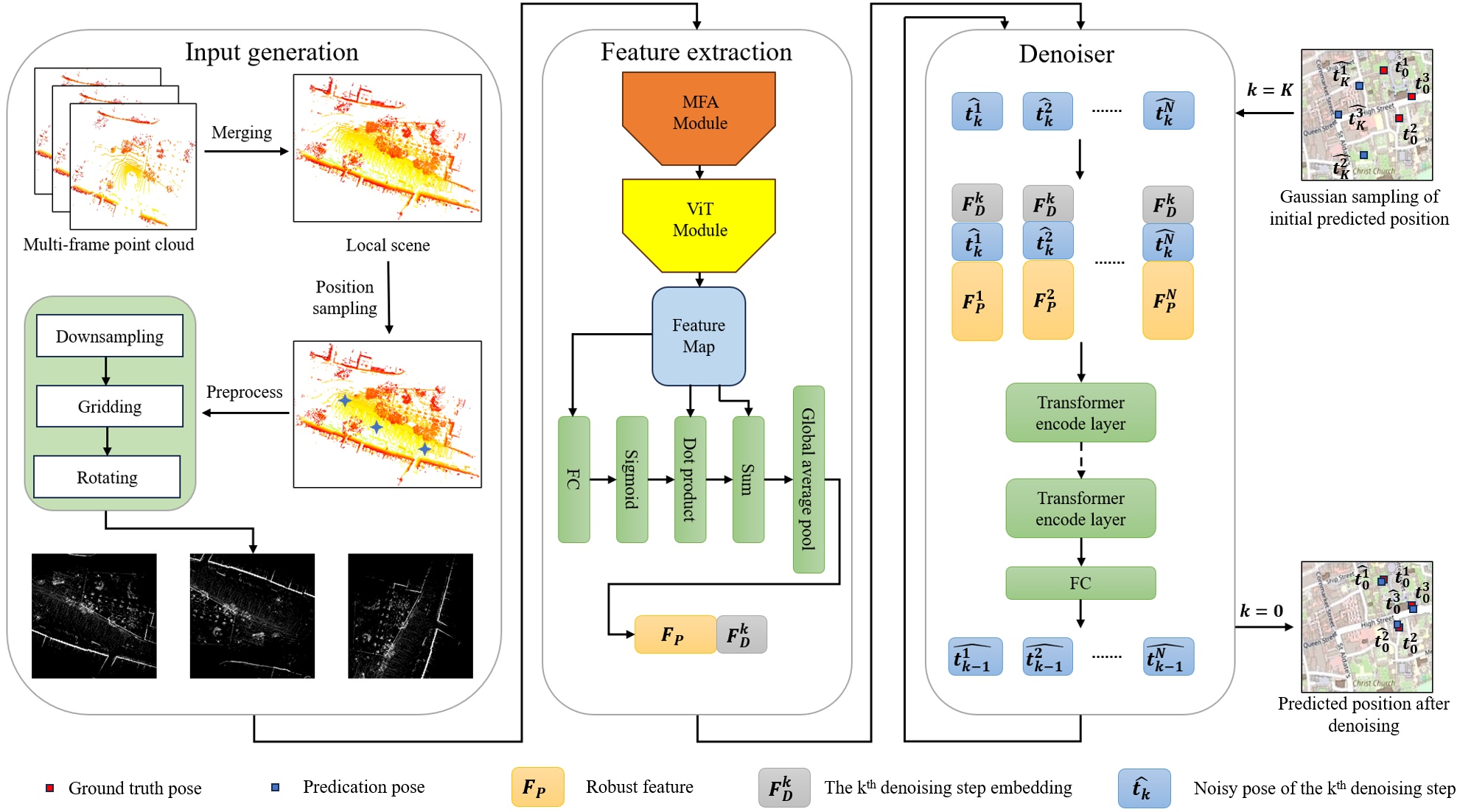}
\caption{The pipeline of the BEVDiffLoc framework. First, it merges continuous $M$ frames of point cloud data with an interval of $S$ frames and generates new BEV images from random positions with random orientations. Then, the MFA module and ViT module are employed to extract robust features $\m F^i_P$, effectively addressing the challenges posed by varying orientations. The pose estimation is modeled as denoising task, ultimately yielding the predicted poses.}
\vspace{-0.2cm}
\label{fig:pipline}
\end{figure*}

\section{Related work}
In this section, we introduce works related to end-to-end localization tasks, which can be broadly categorized into two approaches: absolute pose regression~(APR) and scene coordinate regression~(SCR).

\subsection{Absolute pose regression}
The absolute pose regression (APR) method aims to learn feature descriptions of the scene and then regress its pose. PointLoc~\cite{Wang2022sensors} is the first APR method based on LiDAR, which encodes features using PointNet++~\cite{qi2017nips} followed by a self-attention module. Yu \etal~\cite{Yu2022pr} proposes four different methods for feature extraction. Some studies~\cite{Yu2024tits,Yu2023tits} have explored sequence constraints and achieved significant improvements. HypLiLoc~\cite{wang2023cvpr} fuses multi-modal features in hyperbolic Euclidean space and achieves better results. FlashMix~\cite{goswami2024flashmix} uses a frozen, scene-uncertain backbone to extract local point descriptors, which are aggregated with an MLP mixer to predict the sensor's pose, greatly reducing the training time. DiffLoc~\cite{li2024cvpr} is the state-of-the-art APR method. It uses the basic model and a static object perception pool to learn robust features of range images and employs a diffusion model to extract features as a condition to learn pose restoration. Due to the robustness of LiDAR to changes in lighting conditions, APR based on LiDAR has achieved impressive results in large-scale outdoor scenes.

The works most closely related to our study are HypLiLoc~\cite{wang2023cvpr} and DiffLoc~\cite{li2024cvpr}. While HypLiLoc demonstrates promising results, we identified limitations in its single BEV modality feature extraction strategy. It uses CNN backbone to extract local features of BEV images and directly regresses the pose, without considering the global contextual relationship between different BEV images. To address these shortcomings, we propose a novel approach that enhances feature extraction. Additionally, DiffLoc provides a comprehensive analysis of how the denoising process improves the robustness of APR methods. Building on this insight, we have integrated this method into our framework to further enhance performance. 

\subsection{Scene coordinate regression}

The scene coordinate regression method involves regressing pixels or points to their corresponding 3D coordinates in the scene. This approach primarily relies on random forests~\cite{Brachmann2016cvpr,Cavallari2020tpami} or convolutional neural networks~\cite{Brachmann2018cvpr,Brachmann2022tpmai} to match scenes. SGLoc~\cite{Li2023cvpr} employs the Kabsch algorithm within the RANSAC loop to predict the 3D scene coordinates of each point for pose estimation. Lisa~\cite{yang2024cvpr} leverages diffusion-based distillation in 3D semantic segmentation models to learn multi-scale feature extractors for scene coordinate regression and subsequent pose estimation. Chen \etal~\cite{chen2024cvpr} proposed a method that utilizes a rapidly trained scene coordinate regression model as the scene representation and pre-trains a pose regression network to learn the relationship between scene coordinate prediction and the corresponding camera pose. By leveraging this universal relationship, pose regressors can be trained on hundreds of different scenarios, effectively addressing the issue of limited training data for absolute pose regression models. The limitation of the scene coordinate regression method is that it can only regress the relationship between the query data and the three-dimensional world coordinates. The accuracy of localization is significantly affected by RANSAC. In addition, due to the point cloud stacking in the BEV image, which results in the loss of scene structure details, it is not suitable for the SCR method to perform regression pose.

\section{Our approach}
We propose BEVDiffLoc for end-to-end LiDAR-based localization. As illustrated in~\figref{fig:pipline}, our approach begins by processing the input data, where point clouds are projected into BEV images, which are then stitched into a map, and new BEV images are generated by sampling from random positions to enhance the diversity of the input data, as detailed in \secref{Input Generation}. Next, in the feature extraction module, we extract features with orientation equivariant characteristic, followed by a ViT to enhance feature representation and geometric perception, as described in \secref{Feature Extraction Module}. Finally, the extracted features are fed into a diffusion module to progressively recover the pose, as detailed in \secref{Denoiser}.

\subsection{Input Generation}
\label{Input Generation}
Following existing works~\cite{lu2025tro,luo2024bevplace++}, we assume that ground vehicles travel on a roughly planar surface within a local area. Based on this assumption, we generate BEV images through orthogonal projection and focus on estimating the 3-Degree-of-Freedom~(DoF) pose, including (x, y, yaw). Following Luo~\etal~\cite{luo2021ral}, we construct BEV images using normalized point density. Compared to methods that store the maximum height of pixel midpoints, this approach reduces the influence of sensor orientation on BEV images and enhances their robustness to viewpoint changes~\cite{gupta2024icra}.

Given point cloud $\mathcal{P} = \{ \d p_i \mid i = 1, \dots, N_p \}$ with a total of $N_p$ points. We adopt a right-hand Cartesian coordinate system, where the x-y plane represents the ground plane. Following Luo~\etal~\cite{luo2024bevplace++}, to balance accuracy and computational complexity, we first apply a voxel grid filter with a leaf size of $g$\,m to evenly distribute the points. Subsequently, we define a square window of size [-$L$, $L$]\,m centered at the coordinate origin. The BEV image can be interpreted as an image with dimensions [$\frac{2L}{g}$, $\frac{2L}{g}$]. The pixel value $\mI(u, v)$ in the BEV image is computed as
\begin{equation}
\label{eq1}
	\mI(u, v) = \frac{\min(N_g, N_n)}{N_m},
\end{equation}
where $N_g$ denotes the number of points in the grid at position $(u, v)$, and $N_n$ is the normalization factor, and $N_m$ is the max value of the point cloud density.

To enhance the diversity of training data, we propose a novel data augmentation mechanism. We first construct a local map by stitching continuous $M$ frames of point clouds with an interval of $S$ frames. Subsequently, we randomly sample a new observation position with $[\Delta x, \Delta y]$ offsets and a new observation direction from this local map. Here, $\Delta x$ and $\Delta y$ follow a Gaussian distribution with a mean of $0$ and a variance of $\tau$, while yaw follows a uniform distribution in the range $[0,360]$°. This data augmentation method significantly enriches the scene information from diverse observation positions and angles.

\subsection{Feature Extraction Module}
\label{Feature Extraction Module}
Robust features are crucial for localization. Modern convolutional networks~\cite{luo2024bevplace++} can effectively serve as feature encoders for BEV images, offering translational-equivariance. However, these methods struggle to handle significant rotational view variations. Inspired by recent work~\cite{luo2024bevplace++}, we propose a new feature extraction module to extract robust features with rotational and translational variations. The architecture integrates two core components: a Maximum Feature Aggregation (MFA) module and a Vision Transformer (ViT) module, which are connected in series to form the complete system.

Considering that CNN backbone does not have orientation equivariant characteristic, it cannot handle orientation well. We designed the CNN-based MFA module to address this issue. 
Given an input BEV image $\m I$, we rotate it using a set of angles $\mathcal{R} = \left\{ 0, \frac{2\pi}{N_R}, \dots, \frac{(N_R-1)2\pi}{N_R} \right\}$, and use a weight-shared CNN backbone to extract their local features. Afterwards, rotate the local features at opposite rotation angles and use max-pooling to process them. This method captures the most prominent features at different rotation angles and combines them. When the number of rotations $N_R$ is sufficient, the features extracted by this method will have orientation equivariant characteristic. The maximum feature $\m F_\text{MFA} \in \mathbb{R}^{H \times W \times C}$ is obtained by applying max pooling across the $N_R$ rotational feature maps, given by
\begin{equation}
\label{eq2}
\m F_\text{MFA} = \max_{r \in \mathcal{R}} R_r^{-1} \circ \varphi(R_r \circ \mI),
\end{equation}
where $H$, $W$ and $C$ represent the height, width, and feature channels of the MFA feature map, respectively, and $\varphi$ represents a CNN backbone. In practice, a small $N_R$ is sufficient to achieve satisfactory performance.

Based on the features extracted by MFA, we employ a Vision Transformer (ViT) to enhance feature representation and geometric perception capabilities. In the standard ViT framework, the image is divided into small patches of size $P_H \times P_W$, which are linearly embedded to produce $d$-dimensional visual tokens. To bridge the potential domain gap between feature images and RGB images, we replace the embedding layer with non-linear convolutional stems~\cite{xiao2021neurips}, which have been proven to improve the optimization stability and prediction performance of ViT. 
Then, the output is fed into ViT \cite{dosovitskiy2021iclr} to obtain a feature map $\m F_\text{ViT} \in \mathbb{R}^{M \times C}$, where $M$ and $C$ are the token number (without classification token) and the feature dimension, respectively.

To enhance the learned features and generate corresponding global features, we adopt the following operation. First, we use a fully connected (FC) layer to reduce the channels of ${\m F}_\text{ViT}$ to 1, generating a global feature representation. The second step is to use the sigmoid operator to normalize the global feature representation to the range of $[0,1]$. Then, we perform dot product and addition operations with $\m F_\text{ViT}$ to weight key features and preserve the details of the original features. Finally, we apply global average pooling (GAP) to derive the global feature $\m F_P$, given by

\begin{equation}
\label{eq3}
 \m F_P=\mathrm{GAP}\left(\m F_\text{ViT}+\sigma\left(\mathrm{FC}\left(\m F_\text{ViT}\right)\right)\odot \mF_\text{ViT}\right),
\end{equation}
where $\sigma$ is the sigmoid function and $\odot$ is the dot production.

\subsection{Denoiser for pose refinement}
\label{Denoiser}
Inspired by~\cite{li2024cvpr}, we model LiDAR localization as a denoising process from noisy pose to real pose using a diffusion model based on encoded features $\m F_P$.

During the training stage, we train a diffusion model to learn the underlying distribution of LiDAR poses by recovering the ground truth pose from its corrupted version. In each training iteration $k\in\{1,2,...,K\}$, with a predefined variance schedule $\beta_1, ..., \beta_K$, we introduce noise to the ground truth pose $\d t_0$ following the cumulative noise schedule, obtaining the noisy pose $\d t_k$, calculated as
\begin{equation}
    \label{eq4}
    \begin{split}
q\left(\d t_{k}|\d t_{0}\right)=\mathcal{N}\left(\d t_{k};\sqrt{\overline{\alpha}_{k}}\d t_{0},(1-\overline{\alpha}_{k})\mathbf{I}\right),\\
\d t_{k}=\sqrt{\overline{\alpha}_{k}}\d t_{0}+\sqrt{1-\overline{\alpha}_{k}}\d{\epsilon},\d{\epsilon}\sim\mathcal{N}\left(\mathbf{0},\mathbf{I}\right),
    \end{split}
\end{equation}
where $\alpha_k=1-\beta_k$ and $\overline\alpha_k=\prod_{i=1}^k\alpha_i$. Then, denoiser $\mathcal{D}_{\theta}$ predicts $\epsilon$ to progressively remove noise. We implement the denoiser $\mathcal{D}_{\theta}$ using a transformer $\mathcal{T}$, and thus the denoiser can be calculated as
\begin{equation}
\label{eq5}
\mathcal{D}_\theta\left(\d t_k,k,\mathcal{P}\right)=\mathcal{T}\left(\{\d t_k^i, \m F_D^k, \m F_P^i\}_{i=1}^N\right),
\end{equation}
where $\{\cdot\}$ denotes the concatenation, and the input of $\mathcal{T}$ is the sequence of noisy pose tuples $\d t_k^i$, unique step embedding $\m F_D^k$, and feature embedding $\m F_P^i$. Here, $\m F_D^k$ denotes the $k$-th denoising step generated via a sinusoidal function. 

Following Denoising Diffusion Implicit Model~(DDIM), during the inference stage, we use the trained $\mathcal{D}_{\theta}$ to gradually denoise the pure noise sequence $\hat {\d t}_K$ to obtain the predicted pose $\hat {\d t}_0$. To accelerate the inference process to update the poses, the pose update is performed as 
\begin{equation}
\label{eq6}
\hat{\d t}_{k-1}=\sqrt{\overline{\alpha}_{k-1}}\frac{\hat{\d t}_{k}-\sqrt{1-\overline{\alpha}_{k}}\mathcal{D}_{\theta}}{\sqrt{\overline{\alpha}_{k}}}+\sqrt{1-\overline{\alpha}_{k-1}}\mathcal{D}_{\theta},
\end{equation}
where $\hat {\d t}_{k-1}$ is fed into the denoiser $\mathcal{D}_{\theta}$ for the next step. This iterative denoising is repeated until reaching $\hat {\d t}_0$, which is regarded as the final prediction.

\subsection{Loss function}

Follow DiffLoc~\cite{li2024cvpr}, we employ the following loss to guide the model in predicting noise variable $\epsilon$, given by
\begin{equation}
\label{eq7}
L_1=\|\mathcal{D}_{\theta}\left(\d t_{k},k,\mathcal{P}\right)-\d{\epsilon}\|_{1},
\end{equation}
where the $\d t_k$ and $\d \epsilon$ are defined in~\eqref{eq4}.

\section{Experimental evaluation}
The experiments are conducted to demonstrate the effectiveness of the proposed BEVDiffLoc for LiDAR-based localization. Here, we make three core claims: (i) It achieves state-of-the-art localization accuracy, outperforming existing methods;
(ii) Benefitting from the BEV representation and our designed feature extraction module, our approach significantly enhances the robustness and precision of localization;
(iii) Through the introduction of the specialized data augmentation method tailored for BEV representations, our approach further improves localization performance.

\subsection{Experimental Setup}
\textbf{Datasets.} 
We evaluate the impact of BEVDiffLoc on LiDAR localization using two outdoor benchmark datasets: the Oxford Radar RobotCar~\cite{Barnes2020icra} and the NCLT~\cite{CarlevarisBianco2016ijrr}.
Oxford Radar RobotCar (Oxford)~\cite{Barnes2020icra} is collected in urban scene, with each trajectory spanning approximately 10~\,km. The Oxford dataset includes diverse weather and traffic conditions, making it an ideal choice for comprehensive model evaluation. NCLT~\cite{CarlevarisBianco2016ijrr} dataset offers a variety of scenarios, including outdoor and indoor environments with varying structural complexities, with each trajectory approximately 5.5\,km in length. It consists of data from LiDAR, omnidirectional cameras, and GPS/INS. In our experiments, we use only LiDAR information. 

\textbf{Implementation Details.}
Our BEVDiffLoc is implemented using PyTorch~\cite{paszke2019neurips}. The denoiser is implemented as a Transformer~\cite{vaswani2017neurips} with 8 encoder layers and 4 attention heads for feature aggregation, where the latent embedding dimension is set to 512.
For BEV images, we set the side length of the square window to 50\,m and the resolution of the discretized grid to 0.4\,m. The MFA feature image size is configured as $[32,32,128]$, and the patch size is set to $[4,4]$. For multi-frame data augmentation, we construct a local scene by concatenating $M=5$ consecutive point clouds with an interval of $S=20$ frames. The center position of the scene follows a normal distribution, and the rotation angle is randomly sampled from $[0,360]$°. We define the total denoising steps as $K=100$ and use a batch size of $16$. The input point cloud sequence is a tuple of size $3$ with a spacing of $2$ frames.
We train BEVDiffLoc for $150$ epochs using the AdamW optimizer~\cite{dosovitskiy2021iclr} with a single-cycle cosine annealing strategy~\cite{loshchilov2019iclr} and a linear warm-up. The warm-up epoch and peak learning rate are set to $5$ and $5\times10^{-4}$, respectively. Training is performed on a single RTX 4090 GPU. For the Oxford dataset, we configure the iterative denoising steps to be $10$, while for the NCLT dataset, this value is set to $15$.

\begin{figure*}[h]
\centering
\includegraphics[width=\textwidth]{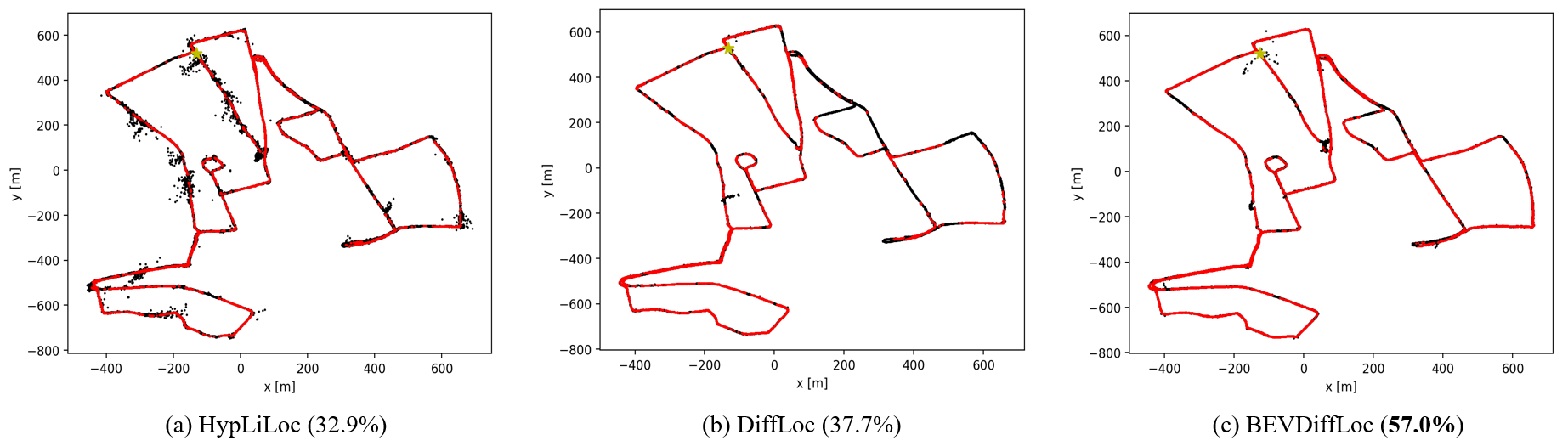}
\caption{Localization results on 17-14-03-00 in the Oxford RobotCar dataset. Successful localizations are marked in red, while failed localizations are marked in black, respectively. The star  denotes the start frame. The caption of each subfigure displays the SR.}
\vspace{-0.2cm}
\label{fig:oxford_result}
\end{figure*}

\begin{figure*}[h]
\centering
\includegraphics[width=\textwidth]{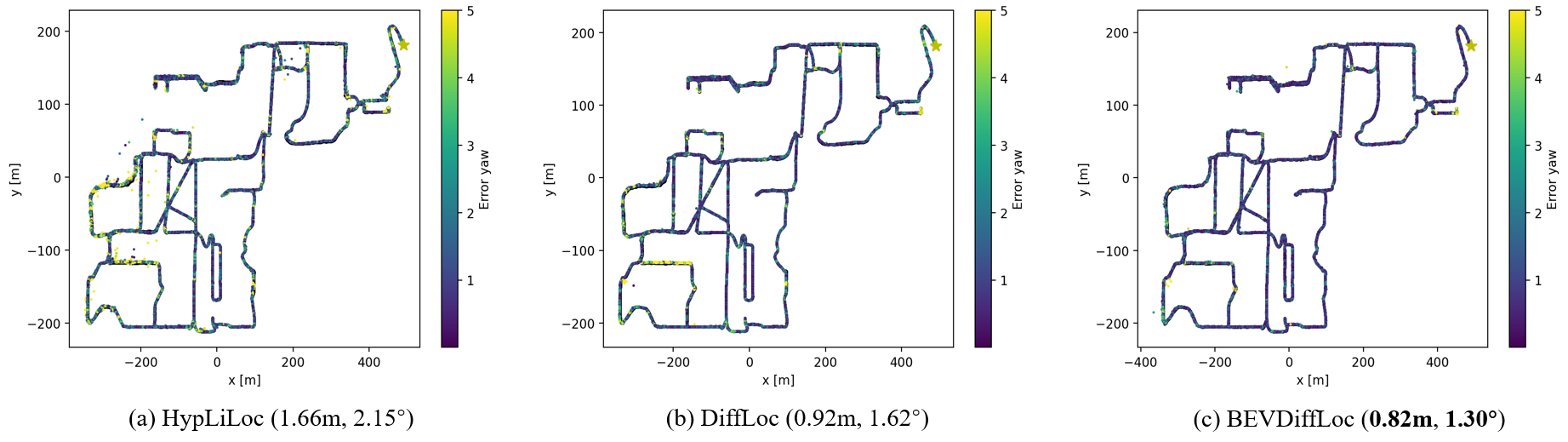}
\caption{Localization results on 2012-02-19 in the NCLT dataset. We use a heatmap with a range of 5 to represent orientation errors. The star denotes the start frame. The caption of each subfigure displays the ${e}_t$ and ${e}_y$.}
\vspace{-0.2cm}
\label{fig:nclt_result}
\end{figure*}
\subsection{Experimental results}
To verify the performance of BEVDiffLoc, we compared it with following state-of-the-art baselines: HypLiLoc~\cite{wang2023cvpr}, DiffLoc~\cite{li2024cvpr}, PosePN~(PPN)~\cite{Yu2022pr}, PosePN++~(PPN++)~\cite{Yu2022pr}, PoseMinkLoc~(PML)~\cite{Yu2022pr} and PoseSOE~(PSOE)~\cite{Yu2022pr}. 
For DiffLoc, we utilize its open-source pre-trained model. For other baselines, since none of them provide pre-trained models, we use their publicly available code and train the models according to the training configurations specified in their papers.
We use mean position error ${e}_t$ and mean orientation error ${e}_y$ to evaluate the 3-DoF estimations of all the baseline methods. Additionally, we compute the localization success rate (SR) under a threshold of 2\,m and 5° for all approaches.

\renewcommand{\arraystretch}{1.5}
\begin{table}[t]
    \centering
    \setlength{\tabcolsep}{1.5pt}
    \caption{Results on Oxford RobotCar Dataset. Best performance is highlighted in bold. The optimal performance is characterized by a lower value of ${e}_t$(m)/${e}_y$(°) and a higher value of SR(\%).}
    \begin{tabular}{lcccccc}
        \toprule
         Sequence & \multicolumn{2}{c}{17-13-26-39} & \multicolumn{2}{c}{17-14-03-00} & \multicolumn{2}{c}{18-14-14-42} \\
         \cmidrule(r){2-3}\cmidrule(r){4-5}\cmidrule(r){6-7}
         &SR(\%)&${e}_t$(m)/${e}_y$(°) &SR(\%)&${e}_t$(m)/${e}_y$(°) &SR(\%)& ${e}_t$(m)/${e}_y$(°) \\
        \hline
        PPN\cite{Yu2022pr} &17.4&17.53/3.37 &14.6&14.10/2.92 &21.2&8.17/1.19  \\
        PPN++\cite{Yu2022pr} &20.3&8.06/1.82 &17.8&6.78/1.67 &28.9&5.77/1.69  \\
        PML\cite{Yu2022pr} &25.2&13.12/2.90 &24.7&10.76/2.75 &38.2&9.36/2.34  \\
        PSOE\cite{Yu2022pr} &24.3&7.96/2.59 &22.2&7.49/2.44 &27.4&6.48/2.01 \\
        HypLiLoc\cite{wang2023cvpr} &42.1&5.56/1.22 &32.9&4.33/0.91 &49.5&2.87/0.72  \\
        DiffLoc\cite{li2024cvpr} &48.4&2.39/\textbf{0.49} &37.7&2.90/\textbf{0.50} &54.3&\textbf{2.14}/\textbf{0.46} \\ \hline
         Ours &\textbf{64.0}&\textbf{2.15}/0.67 &\textbf{57.0}&\textbf{2.80}/0.74 &\textbf{69.4}&3.12/0.75 \\ \bottomrule
    \end{tabular}
    \label{result_oxford}
\end{table}

\renewcommand{\arraystretch}{1.5}
\begin{table}[!h]
    \setlength{\tabcolsep}{1.5pt}
    \centering
    \caption{Results on NCLT Dataset. Best performance is highlighted in bold. The optimal performance is characterized by a lower value of ${e}_t$(m)/${e}_y$(°) and a higher value of SR(\%).}
    \begin{tabular}{lcccccc}
        \toprule
         Sequence&\multicolumn{2}{c}{2012-02-12}&\multicolumn{2}{c}{2012-02-19}&\multicolumn{2}{c}{2012-03-31} \\ 
  \cmidrule(r){2-3}\cmidrule(r){4-5}\cmidrule(r){6-7}
        &SR(\%)&${e}_t$(m)/${e}_y$(°) &SR(\%)&${e}_t$(m)/${e}_y$(°) &SR(\%)& ${e}_t$(m)/${e}_y$(°) \\ \hline
        PPN\cite{Yu2022pr} &43.9&9.26/6.56 &57.7& 6.18/4.11 &57.8&5.55/4.61  \\
        PPN++\cite{Yu2022pr} &52.0&4.47/3.16 &58.0& 2.99/2.01 &56.9&3.24/2.71 \\
        PML\cite{Yu2022pr} &47.4&6.00/4.65 &56.4&4.30/3.10 &54.1& 4.37/3.66 \\
        PSOE\cite{Yu2022pr} &41.0&12.86/7.46 &50.8&5.93/3.81 &49.1&5.08/4.16 \\
        HypLiLoc\cite{wang2023cvpr} &73.2&1.67/2.79 &76.3&1.66/2.15 &74.4&1.61/2.53 \\
        DiffLoc\cite{li2024cvpr} &89.3&0.98/1.81 &89.4&0.92/1.62 &82.7&0.97/1.76  \\ \hline
        Ours &\textbf{93.1}& \textbf{0.95}/\textbf{1.51} &\textbf{94.1}& \textbf{0.82}/\textbf{1.30} &\textbf{93.6}& \textbf{0.88}/\textbf{1.41} \\ \bottomrule
    \end{tabular}
    \label{result_nclt}
\end{table}

\textbf{Results on the Oxford dataset.}
We first evaluate the proposed BEVDiffLoc on the Oxford dataset. As shown in~\tabref{result_oxford}, our method achieves on-par localizaiton accuracy and state-of-the-art localization robustness, with an average improvement of 35.7\% in SR compared to the second-best method, DiffLoc.
\figref{fig:oxford_result} illustrates the predicted trajectories. We color the trajectories based on success rate, providing a more intuitive visualization of the localization robustness of each method. As illustrated, our results are consistent with the findings presented in~\tabref{result_oxford}. Specifically, compared to baseline methods, BEVDiffLoc demonstrates a more stable global localization performance, with fewer outliers.

\textbf{Results on the NCLT dataset.}
Next, we evaluate the proposed DiffLoc on the NCLT dataset. As shown in~\tabref{result_nclt}, our approach demonstrates greater advantages in localization performance, achieving an average improvement of 7.7\% in position and 18.7\% in orientation. We also visualize the trajectories, colored by yaw angle error, predicted by our approach and compare them to DiffLoc and HypLiLoc. As shown in \figref{fig:nclt_result}, our BEVDiffLoc consistently exhibits more stable localization performance with fewer outliers. This further demonstrates the effectiveness of the proposed BEVDiffLoc when dealing with the challenging NCLT dataset.
The superior localization robustness of our method in both the Oxford and NCLT datasets can be attributed to the introduction of BEV representation  and the feature extraction module, which enable the network to directly handle large variations in observation angles, thereby significantly reducing the difficulty of localization in NCLT. Additionally, through data augmentation, our approach greatly enriches the dataset, allowing it to handle a wider variety of scenarios.

\renewcommand{\arraystretch}{1.5}
\begin{table}[t]
    \centering
    \caption{Ablation study on the Oxford and NCLT datasets. 
    }
    \begin{tabular}{cc|cc}
        \toprule
         \multirow{2}*{\textbf{MFA}} & \multirow{2}*{\textbf{Augmentation}} & {Oxford} & {NCLT} \\
         
          &   &SR(\%) ${e}_t$(m)/${e}_y$(°) & SR(\%) ${e}_t$(m)/${e}_y$(°)\\
         \hline
          &  &41.3 5.19/1.34 &40.2 4.56/3.85 \\
         \checkmark &  &53.9 3.28/0.99 &64.1 2.22/2.57 \\ 
         \checkmark & \checkmark &63.5 2.69/0.72 &93.6 0.88/1.41 \\ \bottomrule
    \end{tabular}
    
    \label{Ablation_result}
\end{table}

\subsection{Ablation study}

\textbf{Ablation on MFA module.}
To study the effect of the MFA module, we ablate it and instead use ResNet~\cite{he2016cvpr} to extract BEV features. As shown in~\tabref{Ablation_result}, on the Oxford dataset, MFA module results in a 36.8\% increase in position accuracy and a 26.1\% increase in orientation accuracy. On the NCLT dataset, it achieves a 51.3\% increase in position accuracy and a 33.2\% increase in orientation accuracy.
These results indicate that the orientation equivariance features extracted by the MFA module significantly improve both translation and orientation estimation.

\textbf{Ablation on data augmentation.}
Next, we investigate the proposed data augmentation technique. As shown in~\tabref{Ablation_result}, the proposed data augmentation technique significantly improves performance, achieving an 18.0\% increase in position accuracy and a 27.3\% increase in orientation accuracy on the Oxford dataset. On the NCLT dataset, it results in a 60.4\% increase in position accuracy and a 45.1\% increase in orientation accuracy. Notably, the data augmentation mechanism significantly improves the localization success rate on the NCLT dataset by 46.0\%. This improvement can be attributed to the larger angular variations in the NCLT dataset. Our data augmentation mechanism effectively enables the network to learn and adapt to these variations, thereby enhancing localization robustness.

\subsection{Runtime}
The default LiDAR frame rates for the Oxford and NCLT datasets are 20\,Hz and 10\,Hz, respectively. We tested our method with an Intel(R) Xeon(R) Platinum 8474C, 64 GB of RAM, and a single NVIDIA RTX 4090 GPU. In the Oxford and NCLT datasets, our approach achieves average frame rates of 91\,Hz and 47\,Hz, respectively. This outcome demonstrates that our approach is capable of fulfilling the requirements of online operation.

\section{Conclusion}
In this paper, we explore the BEV representation for addressing the end-to-end localization problem and propose BEVDiffLoc. To fully leverage the advantages of BEV, we stitch multi-frame BEV images into a local map and generate new BEV images by randomly sampling observation positions and angles, significantly enhancing training data diversity and improving the performance. Additionally, we employ a Maximum Feature Aggregation module followed by a Vision Transformer to extract features from BEV images. The feature network enables more direct capture of the scene’s geometric structure, thereby learning more robust and accurate localization features. We formulate the pose estimation process as a denoising process of noisy poses and utilize a diffusion model to output the learned features as poses. Extensive experiments demonstrate the state-of-the-art performance of our method and validate the contribution of each proposed module to the final results.

\bibliographystyle{unsrt}
\bibliography{new}
\end{document}